%% file: main.tex
\documentclass[runningheads]{llncs}
\usepackage{amsmath, amssymb}
\usepackage{xcolor}
\usepackage[T1]{fontenc}
\usepackage{subcaption}
\usepackage{placeins}
\usepackage{xurl}
\usepackage[ruled,vlined]{algorithm2e}

\SetAlFnt{\footnotesize}
\SetAlCapFnt{\footnotesize}
\SetAlCapNameFnt{\footnotesize}
\SetAlgoNlRelativeSize{-2}
\usepackage{graphicx}
\usepackage[ruled,vlined]{algorithm2e}
\usepackage{amsmath}
\usepackage{booktabs}
\usepackage{multirow}
\usepackage{makecell}
\newcommand{\stepidx}[1]{%
  \raisebox{0.2ex}{\scriptsize\textsf{\textbf{#1}}}\,%
}

\newcommand{\stepblock}[2]{%
  \stepidx{#1}\textbf{#2}%
}
\renewcommand{\baselinestretch}{0.93}
\begin{document}
\title{A Unified Complementarity-based Approach for Rigid-Body Manipulation and Motion Prediction}
%
%
\author{Bingkun Huang\inst{1}\and
Xin Ma\inst{1} \and Nilanjan Chakraborty\inst{2} \and
Riddhiman Laha\inst{1,3}}
\authorrunning{Huang et al.}
%
\institute{Technical University of Munich, Munich 80333, Germany\\
\email{\{bingkun.huang,maxin.ma\}@tum.de} \\ \and
Stony Brook University, Stony Brook NY 11733, USA\\
\email{nilanjan.chakraborty@stonybrook.edu}\\ \and
Northeastern University, Boston MA 02115, USA\\
\email{riddhiman.laha@tum.de}}
%
\maketitle              
\begin{abstract}
Robotic manipulation in unstructured environments requires planners to reason jointly about free-space motion and sustained, frictional contact with the environment. Existing (local) planning and simulation frameworks typically separate these regimes or rely on simplified contact representations, particularly when modeling non-convex or distributed contact patches. Such approximations limit the fidelity of contact-mode transitions and hinder the robust execution of contact-rich behaviors in real time. This paper presents a unified discrete-time modeling framework for robotic manipulation that consistently captures both free motion and frictional contact within a single mathematical formalism (Unicomp). Building on complementarity-based rigid-body dynamics, we formulate free-space motion and contact interactions as coupled linear and nonlinear complementarity problems, enabling principled transitions between contact modes without enforcing fixed-contact assumptions. For planar patch contact, we derive a frictional contact model from the maximum power dissipation principle in which the set of admissible contact wrenches is represented by an ellipsoidal limit surface. This representation captures coupled force–moment effects, including torsional friction, while remaining agnostic to the underlying pressure distribution across the contact patch. The resulting formulation yields a discrete-time predictive model that relates generalized velocities and contact wrenches through quadratic constraints and is suitable for real-time optimization-based planning. Experimental results show that the proposed approach enables stable, physically consistent behavior at interactive speeds across tasks, from planar pushing to contact-rich whole-body maneuvers.

\textbf{Video:} \url{https://youtu.be/bJxuN6DvMgs}.

\end{abstract}

\begin{figure}[!t]
    \centering
    \includegraphics[width=0.9\linewidth]{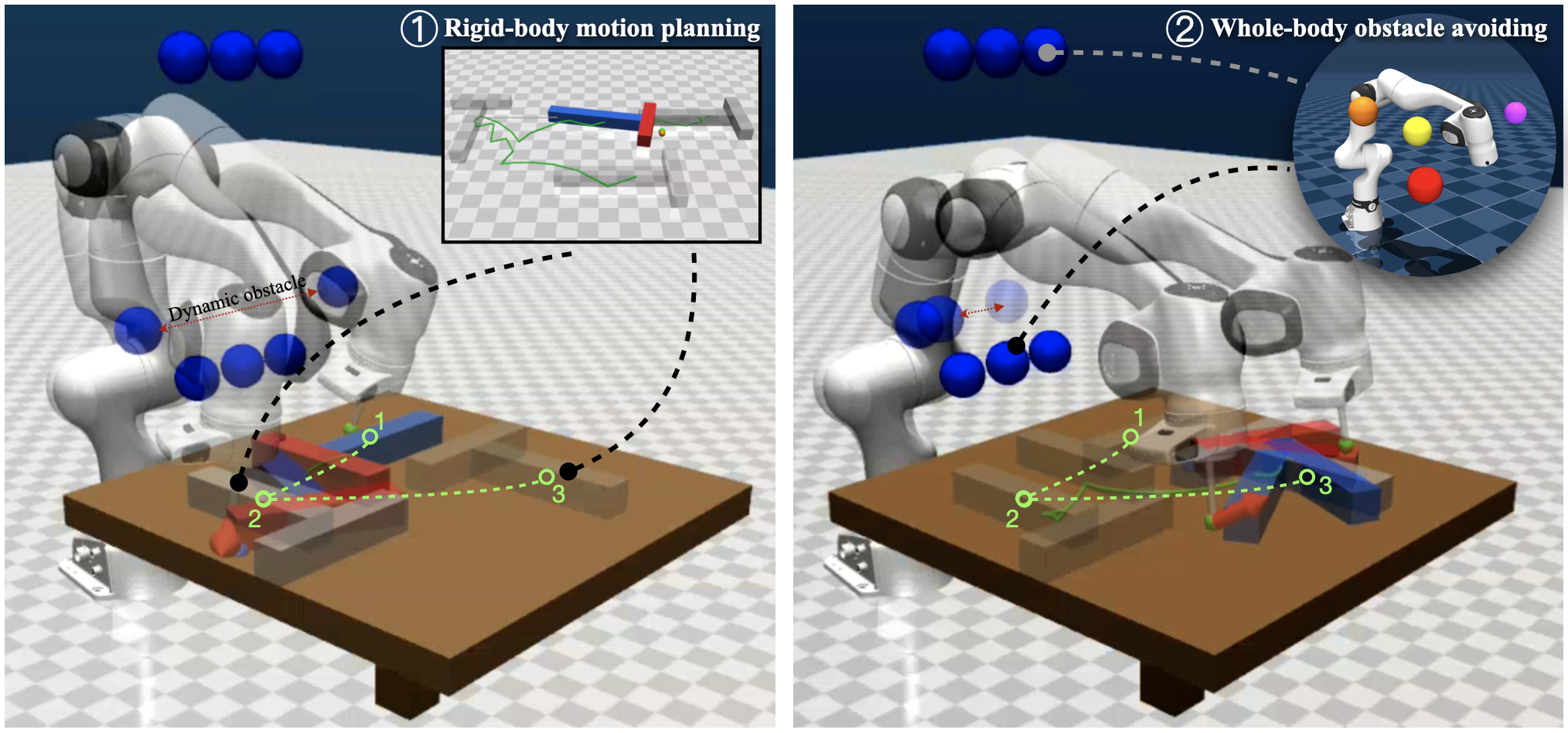}
    \caption{Our real-time software framework allows the system to perform non-prehensile tasks like planar pushing, along with safe whole-body maneuvers.}
    \label{fig:overview}
    \vspace{-0.3cm}
\end{figure}

\vspace{-0.5cm}
\section{Introduction}

Contact implicit planning scenarios (see Fig.~\ref{fig:overview}) where a robotic arm is manipulating an object while exploiting contact with the environment have limitations. They (a) usually assume point contact (or approximate planar patch contact with three-point contact) between the object and the support surface~\cite{selvaggio2021shared}, (b) approximate the friction cone to convert the contact problem into an LCP~\cite{posa2013direct}, and (c) do not include the manipulator motion planning and obstacle avoidance as part of the planning problem. In other words, the contact dynamics between the object and the environment are simplified (at the expense of accuracy)~\cite{carpentier2024compliant}, and it is implicitly assumed that the manipulator does not have any collision avoidance constraints. Thus, the motion planning for the object and the manipulator is decoupled. Motion planning of manipulators is usually done in a kinematic formalism (because robots usually provide more robust position and velocity control interfaces compared to torque control interfaces)~\cite{orthey2023sampling,laha2021point,laha2023predictive}, whereas planning through contact usually uses a dynamic (or quasistatic) modeling paradigm to deal with contact~\cite{horak2019similarities,pang2022easing}. There are no formalisms as well as software implementation that allows one to create/follow a collision-free trajectory for the robot at the kinematic level while computing a trajectory for the objects at the dynamic level, while considering the full dynamic contact model at the contacts. The goal of this paper is to build on recent progress in modeling non-convex planar patch contact as a nonlinear complementarity problem (NCP).
We also incorporate collision avoidance into whole-body manipulator motion planning at the kinematic level using a linear complementarity problem (LCP).
Together, these ideas form a software framework, which we call Unicomp, that uses complementarity constraints to generate and follow collision-free robot trajectories at the kinematic level, while simultaneously computing object trajectories at the dynamic level.  


We provide constructive evidence that such a formulation is feasible and show that it enables real-time execution of complex robotic maneuvers in contact-rich environments. The differential (or dynamic) and mixed complementarity problems are well-known modeling strategies for perfectly rigid bodies in contact~\cite{chakraborty2009optimization,wensing2023optimization,howell2022dojo,tedrake2019drake}. 
Despite these advances, a general and consistent formulation remains lacking for contact patches composed of an arbitrary, potentially infinite, set of contact points. This limitation hinders the integration of contact-rich behaviors within a single real-time computational framework~\cite{jin2024complementarity}. In this work, building upon our prior results, we present a unified formulation of linear and nonlinear complementarity problems (LCPs and NCPs) tailored to complex robotic manipulation scenarios~\cite{yao2025synthesis,muchacho2025unifying,xie2021modeling}. Our approach bridges free-space motion and contact interaction within a common mathematical structure, enabling principled reasoning and efficient computation across diverse contact modes. We consider a friction model derived from the maximum power dissipation principle~\cite{le2024reconciling}, where the set of feasible contact forces and moments is represented by an ellipsoid. Notably, this formulation remains agnostic to the underlying pressure distribution across the contact patch. To summarize, the key contribution of this paper is to present a real-time motion prediction framework for contact-rich tasks that considers both free-space motion with collision avoidance and intermittent contact dynamics within a complementarity-based framework, while rigorously accounting for non-point contact and not approximating friction cone constraints.
    
    

\section{Preliminaries}
\textbf{Notations.} 
The rigid-body configuration is $\boldsymbol{q} = [\mathbf{p_{CoM}}^T,\ \mathbf{Q}^T]^T$, where $\mathbf{p_{CoM}}\in\mathbb{R}^3$ is the center-of-mass position and $\mathbf{Q}\in SO(3)$ is the orientation. The spatial velocity is $\boldsymbol{\nu}=[\boldsymbol{v}^T,\boldsymbol{\omega}^T]^T\in\mathbb{R}^6$. The inertia tensor is $\boldsymbol{M}(\boldsymbol{q})$. The wrench denoted by $\boldsymbol{\lambda}$ and the corresponding impulse is $\boldsymbol{\Lambda}$. We use $\mathbf{R{(.)}}$ to denote the rotation matrix, which is the default from the body frame to the world frame. $\mathbf{a}$ represents the Equivalent Contact Point (ECP).

\subsection{Complementarity problems}
To formalize the unified treatment of free-space motion and contact interaction, we next introduce linear, nonlinear, and mixed complementarity problems (LCPs/NCPs/MCPs), which provide the mathematical backbone for expressing impenetrability, mode-dependent contact constraints, and frictional contact laws within a single solvable framework.

Following \cite{Munson2002Complementarity,trinkle1997dynamic}, we consider the MCP as a unified framework for nonlinear equations and complementarity constraints. Let $\mathbf{F}:\mathbb{R}^n\to\mathbb{R}^n$ be a continuously differentiable mapping, and let the variables be bounded by vectors $\boldsymbol{\ell},\boldsymbol{u}\in(\mathbb{R}\cup\{-\infty,+\infty\})^n$ with $\boldsymbol{\ell}\le \boldsymbol{u}$ (componentwise). The MCP, denoted by $\mathrm{MCP}(\mathbf{F},\boldsymbol{\ell},\boldsymbol{u})$, seeks $\mathbf{z}\in\mathbb{R}^n$ such that for every component $i\in\{1,\dots,n\}$,
\begin{equation}
\begin{cases}
z_i=\ell_i \ \Longrightarrow\ F_i(\mathbf{z})\ge 0,\\[2pt]
\ell_i<z_i<u_i \ \Longrightarrow\ F_i(\mathbf{z})=0,\\[2pt]
z_i=u_i \ \Longrightarrow\ F_i(\mathbf{z})\le 0.
\end{cases}
\label{eq:mcp_munson}
\end{equation}
Equivalently, this can be expressed in the compact bound-complementarity form
\begin{equation}
\ell_i \le z_i \le u_i \ \perp\ F_i(\mathbf{z}), \qquad i=1,\dots,n,
\label{eq:mcp_bound_compact}
\end{equation}
meaning that $F_i(\mathbf{z}) = 0$ whenever $z_i$ lies strictly between its bounds, while $F_i(\mathbf{z}) \geq 0$ at the lower bound and $F_i(\mathbf{z}) \leq 0$ at the upper bound. 

\noindent\textbf{Special cases}.
The MCP formulation encompasses several classical problem classes. In particular, \textbf{(a) Nonlinear equations:} if $\boldsymbol{\ell}=-\infty$ and $\boldsymbol{u}=+\infty$, then \eqref{eq:mcp_munson} reduces to the square system $\mathbf{F}(\mathbf{z})=\mathbf{0}$.\textbf{ (b) NCP:} if $\boldsymbol{\ell}=\mathbf{0}$ and $\boldsymbol{u}=+\infty$, then \eqref{eq:mcp_munson} reduces to the nonlinear complementarity problem $0\le \mathbf{z}\perp \mathbf{F}(\mathbf{z})\ge 0$. and \textbf{(c) LCP:} if, in addition, $\mathbf{F}$ is affine, i.e.,
    \begin{equation}
        \mathbf{F}(\mathbf{z})=\mathbf{M}\mathbf{z}+\mathbf{q},
    \end{equation}
    with $\mathbf{M}\in\mathbb{R}^{n\times n}$ and $\mathbf{q}\in\mathbb{R}^n$, then \eqref{eq:mcp_munson} further reduces to the linear complementarity problem~\cite{anitescu1997formulating}
    \begin{equation}
        \mathbf{0}\le \mathbf{z}\ \perp\ \mathbf{M}\mathbf{z}+\mathbf{q}\ge \mathbf{0}.
        \label{eq:lcp_as_mcp}
    \end{equation}

\subsection{Dynamics of Bodies in Contact}

\begin{figure}[!h]
  \centering
  \begin{minipage}[c]{0.30\linewidth}
    \centering
    \includegraphics[width=\linewidth]{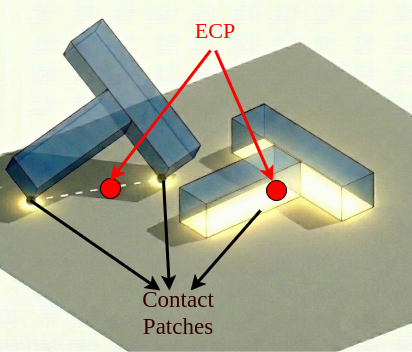}
    \captionof{figure}{ECP visualized}
    \label{fig:ECP_illustration}
  \end{minipage}\hfill
  \begin{minipage}[c]{0.62\linewidth}
    \small
    We adopt a modeling framework similar to our previous work~\cite{ChakrabortyBAT14,Xie2016SIMPAR,xie2019dynamicmodelplanarsliding}, which represents a non-point (line/surface) contact with an equivalent contact wrench acting at the Equivalent Contact Point (ECP). Using the
concept of ECP, we present a principled method for
simulating intermittent contact with convex contact patches.
    The contact wrench over the contact patch between rigid bodies can be equivalently expressed as a wrench applied at ECP, illustrated in Fig.~\ref{fig:ECP_illustration}.
  \end{minipage}
\end{figure}



The Newton-Euler Equation of motion of rigid body are:
\begin{equation}
\mathbf{M}(\mathbf{q})\dot{\boldsymbol{\nu}}
=
\mathbf{W}_n \lambda_n + \mathbf{W}_t \lambda_t + \mathbf{W}_o \lambda_o + \mathbf{W}_r \lambda_r
+ \boldsymbol{\lambda}_{\mathrm{app}} + \boldsymbol{\lambda}_{\mathrm{vp}},
\label{eq:newton_euler_general_ecp}
\end{equation}
where $\lambda_n$ is the normal contact force magnitude, $\lambda_t,\lambda_o$ are tangential force components in the contact frame, $\lambda_r$ is the torsional (about-normal) friction moment, $\lambda_{app}$ is the vector of external forces and moments and $\lambda_{vp}$ is the centripetal and Coriolis forces; $\mathbf{W}_{\{\cdot\}}$ map contact forces/moments from the ECP to CoM. 

The normal constraint is enforced by complementarity between the normal force and the gap function at the ECP,
\begin{equation}
0 \le \lambda_n \ \perp\ \Psi(\mathbf{a_1},\mathbf{a_2}) \ge 0,
\label{eq:normal_comp}
\end{equation}
where $\Psi(\mathbf{a_1},\mathbf{a_2})$ denotes the signed normal separation between the bodies evaluated at the ECP $\mathbf{a_1}, \mathbf{a}_2$.

We assume that the tangential forces and torsional moment at the ECP are selected according to the maximum power dissipation principle under an ellipsoidal admissible set \cite{GOYAL1991307_P1,GOYAL1991331_P2}. Let $v_t, v_o$ be the corresponding tangential components of the relative velocity at the ECP, and let $v_r$ be the relative angular velocity about the contact normal. We collect them as $\boldsymbol{v}_f \triangleq [v_t, v_o, v_r]^T $, and define the friction wrench $\boldsymbol{\lambda}_f \triangleq [\,\lambda_t,\ \lambda_o,\ \lambda_r\,]^T$, which can be obtained as the solution of
\begin{equation}
\begin{aligned}
\max_{\lambda_t,\lambda_o,\lambda_r}\quad &-\left(v_t\lambda_t + v_o\lambda_o + v_r\lambda_r\right)\\
\text{s.t.}\quad &
\left(\frac{\lambda_t}{e_t}\right)^2+\left(\frac{\lambda_o}{e_o}\right)^2+\left(\frac{\lambda_r}{e_r}\right)^2-\mu^2\lambda_n^2 \le 0,
\end{aligned}
\label{eq:fric_mpd}
\end{equation}
where $\mu$ is the coefficient of friction, and $e_t,e_o,e_r>0$ are positive constants defining the friction ellipsoid. Using the Fritz--John optimality conditions, \eqref{eq:fric_mpd} admits an equivalent NCPs representation by \eqref{eq:fj1}--\eqref{eq:fj4}, where $\sigma$ is the Lagrange multiplier associated with the ellipsoidal constraint.
\begin{align}
0 &= e_t^{2}\mu\lambda_n v_t + \lambda_t \sigma, \label{eq:fj1}\\
0 &= e_o^{2}\mu\lambda_n v_o + \lambda_o \sigma, \label{eq:fj2}\\
0 &= e_r^{2}\mu\lambda_n v_r + \lambda_r \sigma, \label{eq:fj3}\\
0 &\le \mu^{2}\lambda_n^{2} - \lambda_t^{2}/e_t^{2} - \lambda_o^{2}/e_o^{2} - \lambda_r^{2}/e_r^{2}\ \perp\ \sigma \ge 0, \label{eq:fj4}
\end{align}
\noindent\textbf{Modes and Switching}. Together with unilateral contact complementarity~\cite{moreau1988unilateral}, these conditions induce three regimes (breaking, sticking, and sliding contact): \textbf{(1) Breaking contact}. If $\Psi(\mathbf{a_1},\mathbf{a_2}) > 0$, then $\lambda_n =0$ and no friction wrench transmitted,i.e., $\boldsymbol{\lambda}_f =0$.
\textbf{(2) Sticking contact}. If $\Psi(\mathbf{a_1},\mathbf{a_2}) = 0$ and $\boldsymbol{v}_f = 0$, then the admissible-set constraint can be inactive, meaning the slack in \eqref{eq:fj4} is strictly positive, and hence $\sigma = 0$ while $\boldsymbol{\lambda}_f$ lie inside the ellipsoid. 
\textbf{(3) Sliding contact}. If $\Psi(\mathbf{a_1},\mathbf{a_2}) = 0$ and $ \left \|\boldsymbol{v}_f \right \| > 0$, maximum dissipation implies the admissible-set constraint is active, therefore $\sigma > 0$, and $\boldsymbol{\lambda}_f$ lies on the ellipsoid boundary. 

Therefore, the mode transitions are governed by the complementarity structure: (i) breaking contact is triggered by $(\Psi, \lambda_n)$, (ii) sticking-sliding switch is governed by activation of the limit surface constraint (inactive: $\sigma = 0 $ for sticking; active: $\sigma >0 $ for sliding).

\section{Problem Formulation}
\label{sec:problem}
Our goal is to provide a discrete-time model that (i) enforces non-penetration during free motion and (ii) predicts frictional contact dynamics under a planar patch
contact, enabling online planning via receding-horizon prediction.

\textbf{System, State, and Inputs.}
\label{sec:state}
We consider a rigid-body system evolving in discrete time with configuration
$\boldsymbol{q}^u \in \mathrm{SE}(3)$ and generalized velocity $\boldsymbol{\nu}^u$.
Contact interactions are represented by a contact wrench $\lambda^u$,
which is constrained to lie within the friction cone
$
\label{eq:friction_cone}
\mathcal{F}
=
\left\{
\boldsymbol{\lambda}^u \;\middle|\;
\boldsymbol{\lambda}^{u\top}
\boldsymbol{W}^{-1}
\boldsymbol{\lambda}^u
\le
\mu^2 \lambda_{n,u}^2
\right\}.
$

The system configuration is updated according to
$
\boldsymbol{q}^{u+1}
=
\boldsymbol{q}^u
\oplus
\Delta t\,\boldsymbol{\nu}^{u+1},
$

where $\oplus$ is integration on SE(3).

\textbf{Optimization Problem.}
\label{sec:obti}
Let the system be at time $t_u$ with state $(\boldsymbol{q}^u,\boldsymbol{\nu}^{u})$ and time step $\Delta t$. We seek the next-step velocity and contact wrench $(\boldsymbol{\nu}^{u+1},\boldsymbol{\lambda}^u)$ that are dynamically feasible, enabling real-time, collision-safe whole-body maneuvers. The following optimization problem can be formulated:
\begin{equation}
\label{eq:single_step_contact}
\begin{aligned}
\min_{\boldsymbol{\nu}^{u+1},\,\boldsymbol{\lambda}^u}
\quad &
\bf{1} \\
\text{s.t.}\quad &
\boldsymbol{M}(\boldsymbol{q}^u)
\left(
\boldsymbol{\nu}^{u+1} - \boldsymbol{\nu}^u
\right)
=
\Delta t
\left(
\boldsymbol{\tau}_{\mathrm{ext},u}
- \boldsymbol{h}(\boldsymbol{q}^u,\boldsymbol{\nu}^{u+1})
+ \boldsymbol{J}^\top(\boldsymbol{q}^u)\boldsymbol{\lambda}^u
\right), \\
&
\boldsymbol{\phi}(\boldsymbol{q}^u)
+
\Delta t\,\boldsymbol{J}_n(\boldsymbol{q}^u)\boldsymbol{\nu}^{u+1}
\ge \mathbf{0}, \\
&
\boldsymbol{\lambda}_{n}^{u} \ge \mathbf{0}, \\
&
\boldsymbol{\lambda}^u \in \mathcal{F}, \\
&
\boldsymbol{J}^u(\boldsymbol{q}^u)\boldsymbol{\nu}^{u+1}
\;\perp\;
\boldsymbol{\lambda}^{u}_t.
\end{aligned}
\end{equation}

The constraints enforce discrete-time rigid-body dynamics, non-penetration,
unilateral contact, friction cone admissibility, and complementarity between
tangential velocity and friction forces.



\section{Unified Complementarity-Based Motion Model}

\textbf{Complementarity Problems of Single Rigid Body.}
Building on the ECP-based contact dynamics reviewed in the previous section, we now present our time-stepping formulation based on the forward Euler method \cite{Iserles_2008}, with all quantities expressed in an inertial (non-rotating) world frame. Let $t_u$ denote the current time and $h$ be the duration of each time step, the superscript $u$ represents the beginning of the current time, and $u+1$ is the end of the current time. Therefore $\dot{\boldsymbol{\nu}} \approx (\boldsymbol{\nu}^{u+1} - \boldsymbol{\nu}^{u})/h$,  $\dot{\boldsymbol{q}} \approx (\boldsymbol{q}^{u+1} - \boldsymbol{q}^{u})/h$, and the force has the impulse format $\Lambda_{(.)} = h\lambda_{(.)}$. If we ignore contact interactions and consider only external impulse, the Newton-Euler equation in discrete time becomes:
\begin{equation}
\boldsymbol{\nu}^{u+1}_{\mathrm{free}}
=
\boldsymbol{\nu}^{u}
+
\boldsymbol{M}(\boldsymbol{q}^u)^{-1}\mathbf{\Lambda}_{app},
\label{eq:free_velocity}
\end{equation}
where $\mathbf{\Lambda}_{app}$ is the external forces and moments applied on CoM. For a free rigid body with configuration $\boldsymbol{q}^u$ the generalized inertia matrix has a block-diagonal form~\cite{siciliano2009robotics}
\begin{equation}
    \boldsymbol{M}(\boldsymbol{q}^u) = 
    \begin{bmatrix}
        m\mathbf{I}_3 & \mathbf{0} \\
        \mathbf{0}    & \mathbf{I}_w(\mathbf{Q}^u)
    \end{bmatrix},
    \qquad
    \mathbf{I}_w(\mathbf{Q}^u) = \mathbf{R}(\mathbf{Q})\mathbf{I}_b\mathbf{R}(\mathbf{Q})^{T}
\label{mass_matrix}
\end{equation}
where $\mathbf{I}_b$ is the inertia tensor in body frame about CoM, and $\mathbf{I}_w$ is the projection of $\mathbf{I}_b$ from body frame to the world frame. And its inverse is therefore:

\begin{equation}
    \boldsymbol{M}(\boldsymbol{q}^u)^{-1} = 
    \begin{bmatrix}
        \frac{1}{m}\mathbf{I}_3 & \mathbf{0} \\
        \mathbf{0}              &  \mathbf{I}_w(\mathbf{Q}^u)^{-1}
    \end{bmatrix},
\end{equation}

\textbf{Dynamics of bodies with contact}. At each time step, we require an initial estimate of the ECP on the body to warm up the solver. We pre-sample a set of vertices $\{\mathbf{x}_i\}_{i=1}^{N}$ that approximate the rigid body convex hull in the body frame (e.g. the box corners of all constituent geom). Given the configuration $\boldsymbol{q}^u$, we transform these vertices into world coordinates as $\mathbf{x}_i^w(\boldsymbol{q}^u) = \mathbf{p}^u + \mathbf{R}(\mathbf{Q}^u)\,\mathbf{x}_i $. We then select the vertex along the contact plane $\mathbf{n}_\pi$,
\begin{equation}
    i^\star \in \arg \min_i (\mathbf{n}_\pi^T\mathbf{x}_i^w(\mathbf{q}^u) - \mathbf{d}_\pi), \qquad \mathbf{a}_0^u \triangleq \mathbf{x}_{i ^ \star}^w(\boldsymbol{q}^u)
    \label{initial_ECP}
\end{equation}
where $\mathbf{d}_\pi$ is the plane offset, if the contact plane is ground, then the formulation will be $i^\star \in \arg \min_i \mathbf{n}_z^T\mathbf{x}_i^w(\mathbf{q}^u) $, and $\mathbf{n}_z^T = [0,0,1]$. This initialization is consistent with the fact that, under the gravity-driven contact with ground, the support region is contained in the set of lowest points of body. In the subsequent contact solve progress, ECP $\mathbf{a}^u$ is treated as a decision point constrained to lie within the convex hull of body. We represent the convex hull of body by half-spaces in the world frame:
\begin{equation}
    \mathbf{A}_w(\mathbf{q}^u)\mathbf{a}^u\leq \mathbf{b}_w(\mathbf{q})
    \label{half-space-constraints}
\end{equation}
where $\mathbf{A}_w(\mathbf{q}^u) \in \mathbb{R}^{m\times3}$ define $m$ half-spaces constraints by each row is the norm vector of a plane of convex body. $\mathbf{b}_w(\mathbf{q}) \in \mathbb{R}^m$ is offset for each plane. We incorporate \eqref{half-space-constraints} via KKT complementarity conditions by introduce dual multipliers $\boldsymbol{l} \geq \mathbf{0}$:
\begin{align}
    \mathbf{0} & =\nabla_\mathbf{a}\phi(\mathbf{a}^u) +  \mathbf{A}_w(\mathbf{q}^u)^T\boldsymbol{l}^u \label{ECP_Complentary_1} \\
    \mathbf{0} & \leq \mathbf{b}_w(\mathbf{q}^u) - \mathbf{A}_w(\mathbf{q}^u)\mathbf{a}^u \perp \boldsymbol{l}^u \geq \mathbf{0}, \label{ECP_Complentary_2} 
\end{align}
where $\phi(\mathbf{a}^u)$ is the tie-break potential function, which we use to guide the ECP solving.
\begin{equation}
    \phi(\mathbf{a}^u) = \frac{\rho}{2} \left \| \mathbf{a}^u_{xy} - \mathbf{a}^u_{0,xy} \right \|_2^2 + \mathbf{n}_\pi^T\mathbf{a}^u - d_\pi
    \label{tie_break_plane}
\end{equation}
In particular for ground plane, \eqref{ECP_Complentary_1} implies
\begin{equation}
    \nabla_\mathbf{a}\phi(\mathbf{a}^u) = 
    \begin{bmatrix}
        \rho(a_x^u - a_{0,x}^u) \\
        \rho(a_y^u - a_{0,y}^u) \\
        1
    \end{bmatrix},
    \label{gradient_tie_break}
\end{equation}
which breaks ties by weakly attracting the tangential ECP coordinates toward $\mathbf{a}^u_{xy}$ with coefficient $\rho$, while favoring the lowest point along the ground normal.  

Let $\mathbf{\Lambda}_c$ denote the net contact (wrench) impulse acting on the rigid body over time steps. The post impact spatial velocity is obtained by
\begin{equation}
\boldsymbol{\nu}^{u+1}
=
\boldsymbol{\nu}^{u}_{free}
+
\boldsymbol{M}(\boldsymbol{q}^u)^{-1}\mathbf{\Lambda}_{c}.
\label{eq:post_velocity}
\end{equation}
Given the post-impact spatial velocity, we can update the configuration by integration the kinematics over the time step.
\begin{equation}
    \mathbf{q}^{u+1} = \mathbf{q}^{u} + h\mathbf{G}(\mathbf{q}^{u})\boldsymbol{\nu}^{u+1} 
    \label{eq:config_update}
\end{equation}
where $\mathbf{G}(\mathbf{q}^{u}) \in \mathbb{R}^{6 \times 7}$ maps the generalized velocity $\boldsymbol{\nu}$ to the time derivative of the position and orientation, $\dot{\mathbf{q}} = \mathbf{G}(\mathbf{q})\boldsymbol{\nu}$. 

In our ECP formulation, the contact impulse is parameterized at $\mathbf{a}^u$ by two tangential  impulses $(\Lambda_t, \Lambda_o)$, and a torsional impulse $\Lambda_r$ about contact normal. Let $\mathbf{n}, \mathbf{t}, \mathbf{o} \in \mathbb{R}^3$ be the contact normal and two orthonormal tangents, and define the moment arm $\mathbf{r}^u = \mathbf{a}^u - \mathbf{p}_{CoM}^u$. Then 
\begin{equation}
    \mathbf{\Lambda}_c = 
    \begin{bmatrix}
        \mathbf{n}\Lambda_n^{u+1} + \mathbf{t}\Lambda_t^{u+1} + \mathbf{o}\Lambda_o^{u+1} \\
        \mathbf{r}^u\times(\mathbf{n}\Lambda_n^{u+1} + \mathbf{t}\Lambda_t^{u+1} + \mathbf{o}\Lambda_o^{u+1}) + \mathbf{n}\Lambda_r^{u+1}
    \end{bmatrix}.
    \label{contact_wrench}
\end{equation}

Let $\Psi(\mathbf{q}^u, (\mathbf{a_1^u}, \mathbf{a_2^u}) )$ denotes the signed normal gap between ECP pair (positive when separated). We solve for one ECP using the complementarity equations above, while the second ECP is constructed by projecting the previous ECP onto the corresponding contact plane. Let $\nu_n^{u+1}$ be the normal relative velocity, we enforce the discrete-time non-penetration via complementarity condition

\begin{equation}
    0 \leq \Lambda_n \perp \Psi(\mathbf{q}^u, (\mathbf{a_1^u}, \mathbf{a_2^u})) + h\nu_n^{u+1} + \varepsilon\Lambda_n \geq 0 ,
    \label{non_penetration_disc}
\end{equation}
where $\varepsilon > 0$ is a small regulation parameter. 

Following the previous section, we employ the maximum power dissipation principle with ellipsoidal friction admissible set. Let $\mathbf{\Lambda}_f = [\Lambda_t, \Lambda_o, \Lambda_r]^T$ be the tangential and torsional friction impulse vector, and let $\boldsymbol{\nu}_f = [\nu_t, \nu_o, \nu_r]^T$ be the corresponding slip, twist velocity at ECP expressed in contact frame. The friction impulse is selected by     
\begin{equation}
\begin{aligned}
\max_{\mathbf{\Lambda}_f^{u+1}} \quad &   - (\mathbf{\Lambda}_f^{u+1})^T \boldsymbol{\nu}_f^{u+1}  \\
\text{s.t.}\quad &
\left(\frac{\Lambda_t^{u+1}}{e_t}\right)^2+\left(\frac{\Lambda_o^{u+1}}{e_o}\right)^2+\left(\frac{\Lambda_r^{u+1}}{e_r}\right)^2-\mu^2(\Lambda_n^{u+1})^2 \le 0.
\end{aligned}
\label{eq:fric_impulse}
\end{equation}
Also, using the Fritz-John optimality conditions, \eqref{eq:fric_impulse} can be reformulated into an equivalent complementarity representation: 
\begin{align}
0 &= e_t^{2}\mu\Lambda_n^{u+1} \nu_t^{u+1} + \Lambda_t^{u+1} \sigma^{u+1}, \label{eq:fj_impulse_1}\\
0 &= e_o^{2}\mu\Lambda_n^{u+1} \nu_o^{u+1} + \Lambda_o^{u+1} \sigma^{u+1}, \label{eq:fj_impulse_2}\\
0 &= e_r^{2}\mu\Lambda_n^{u+1} \nu_r^{u+1} + \Lambda_r^{u+1} \sigma^{u+1}, \label{eq:fj_impulse_3}\\
0 &\le \mu^{2}(\Lambda_n^{u+1})^{2} - (\Lambda_t^{u+1})^{2}/e_t^{2} - (\Lambda_o^{u+1})^{2}/e_o^{2} - (\Lambda_r^{u+1})^{2}/e_r^{2}\ \perp\ \sigma^{u+1} \ge 0.\label{eq:fj_impulse_4}
\end{align}

Equation \eqref{eq:free_velocity} - \eqref{eq:fj_impulse_4} define a MCP in unknowns
\begin{equation}
    \mathbf{z} = [\boldsymbol{\nu}^{u+1}, \Lambda_t^{u+1}, \Lambda_o^{u+1}, \Lambda_r^{u+1}, \mathbf{a}^{u+1}, \Lambda_n^{u+1}, \boldsymbol{l}^{u}, \sigma^{u+1}]^T.
    \label{unkowns}
\end{equation}

\subsection{Complementarity Problems between Multiple Rigid Bodies}
For a contact between two rigid bodies $j\in\{1,2\}$, we retain the same ECP-based impulse model and the same complementarity structure as in the single-body case. Only the following equations are modified. Let $\boldsymbol{\lambda}_l^{u+1}\triangleq \mathbf{n}\Lambda_n^{u+1}+\mathbf{t}\Lambda_t^{u+1}+\mathbf{o}\Lambda_o^{u+1}$. With $\mathbf{r}_j^u\triangleq \mathbf{a}_j^u-\mathbf{p}_{\mathrm{CoM},j}^u$, the wrench impulses applied to the two bodies satisfy
\begin{equation}
\mathbf{\Lambda}_{c}^{(1)} =
\begin{bmatrix}
\boldsymbol{\lambda}_l^{u+1}\\
\mathbf{r}_1^u \times \boldsymbol{\lambda}_l^{u+1} + \mathbf{n}\Lambda_r^{u+1}
\end{bmatrix},
\qquad
\mathbf{\Lambda}_{c}^{(2)} =
\begin{bmatrix}
-\boldsymbol{\lambda}_l^{u+1}\\
\mathbf{r}_2^u \times (-\boldsymbol{\lambda}_l^{u+1}) - \mathbf{n}\Lambda_r^{u+1}
\end{bmatrix}.
\label{eq:two_body_action_reaction}
\end{equation}
Accordingly, the post-impact velocities are updated by
\begin{equation}
\boldsymbol{\nu}_{j}^{u+1}
=
\boldsymbol{\nu}_{j,\mathrm{free}}^{u+1}
+
\mathbf{M}_j(\mathbf{q}_j^u)^{-1}\mathbf{\Lambda}_{c}^{(j)},
\qquad j\in\{1,2\}.
\label{eq:two_body_post_velocity}
\end{equation}

The slip/twist velocities used in the friction model are computed from the \emph{relative} motion at the ECP pair:
\begin{equation}
\mathbf{v}_{\mathrm{rel}}^{u+1}
=
\Big[\mathbf{I}_3\ \ -[\mathbf{r}_1^u]_\times\Big]\boldsymbol{\nu}_1^{u+1}
-
\Big[\mathbf{I}_3\ \ -[\mathbf{r}_2^u]_\times\Big]\boldsymbol{\nu}_2^{u+1},
\label{eq:two_body_vrel_nu}
\end{equation}
and the contact-frame components are
\begin{equation}
\begin{aligned}
\nu_n^{u+1}&=\mathbf{n}^T\mathbf{v}_{\mathrm{rel}}^{u+1},\qquad
\nu_t^{u+1}=\mathbf{t}^T\mathbf{v}_{\mathrm{rel}}^{u+1},\qquad
\nu_o^{u+1}=\mathbf{o}^T\mathbf{v}_{\mathrm{rel}}^{u+1},\\
\nu_r^{u+1}&=\mathbf{n}^T\mathbf{S}_\omega\!\left(\boldsymbol{\nu}_1^{u+1}-\boldsymbol{\nu}_2^{u+1}\right).
\end{aligned}
\label{eq:two_body_contact_components_unified}
\end{equation}
where $\mathbf{S}_\omega \triangleq \big[\mathbf{0}_{3\times 3}\ \ \mathbf{I}_3\big]$ extracts the angular-velocity part from the spatial velocity $\boldsymbol{\nu}=[\mathbf{v}^T,\boldsymbol{\omega}^T]^T$. These components replace the single-body counterparts in \eqref{eq:fj_impulse_1}--\eqref{eq:fj_impulse_4} without further modification.

The discrete non-penetration condition uses a two-body gap function:
\begin{equation}
0 \leq \Lambda_n^{u+1} \ \perp\
\Psi(\mathbf{q}_1^u,\mathbf{q}_2^u,\mathbf{a}_1^u,\mathbf{a}_2^u)
+ h\nu_n^{u+1} + \varepsilon\Lambda_n^{u+1}
\geq 0,
\label{eq:two_body_nonpen}
\end{equation}
while the ECP selection for one body remains governed by \eqref{ECP_Complentary_1}--\eqref{ECP_Complentary_2}, and the paired ECP on the other body is obtained by projection onto its local supporting plane as described above.

In Alg. \ref{alg:mb_contact_interactions_loop}, we adopt an implicit time-stepping formulation for multi-rigid-body contact: at each step we (i) perform contact detection, (ii) if contacts exist, compute contact impulses by solving a MCP, and (iii) update all poses and velocities simultaneously. 


\subsection{Complementarity Problems in Whole-body Collision Avoidance}
We cast whole-body collision avoidance as a complementarity-based projection on the same discrete-time grid used throughout this paper. Let $\boldsymbol{\theta}^u\in\mathbb{R}^{n}$ denote the joint configuration and let
$
\boldsymbol{\theta}^{u+1} =\boldsymbol{\theta}^{u} + h\dot{\boldsymbol{\theta}}^{u}
$
be the discrete-time update. Equivalently, define the joint increment
$
\Delta\boldsymbol{\theta}^{u} = \boldsymbol{\theta}^{u+1}-\boldsymbol{\theta}^{u} = h\dot{\boldsymbol{\theta}}^{u}.
$
We approximate the robot's whole-body geometry by a set of spheres rigidly attached to links. For each robot sphere $s\in\{1,\dots,N_s\}$ with radius $r_s>0$, let $\mathbf{p}_s(\boldsymbol{\theta}^u)\in\mathbb{R}^3$ be its world-frame center and let
$\mathbf{J}_s(\boldsymbol{\theta}^u)\in\mathbb{R}^{3\times n}$ be the translational Jacobian such that
$
\Delta\mathbf{p}_s^{u} \approx \mathbf{J}_s(\boldsymbol{\theta}^u)\Delta\boldsymbol{\theta}^{u}.
\label{eq:delta_ps_linearization}
$

Obstacles are represented by spheres with centers $\mathbf{p}_o^u\in\mathbb{R}^3$ and radii $r_o>0$. We also allow moving obstacles with estimated translational velocity $\mathbf{v}_o^u$, so that their discrete displacement over one step is approximated by $h\,\mathbf{v}_o^u$. For each robot--obstacle sphere pair $(s,o)$, define the signed gap distance:
\begin{equation}
\Psi_{so}(\boldsymbol{\theta}^u)
= 
\|\mathbf{p}_s(\boldsymbol{\theta}^u)-\mathbf{p}_o^u\|
-\left(r_s+r_o\right)-d_{\min},
\label{eq:clearance_so}
\end{equation}
where $d_{\min}>0$ is a prescribed safety margin. Let
\begin{equation}
\mathbf{n}_{so}^u \;\triangleq\;
\frac{\mathbf{p}_s(\boldsymbol{\theta}^u)-\mathbf{p}_o^u}{\|\mathbf{p}_s(\boldsymbol{\theta}^u)-\mathbf{p}_o^u\|}
\label{eq:normal_so}
\end{equation}
be the outward unit normal from the obstacle sphere to the robot sphere. We impose a one-step separation condition in discrete time
\begin{equation}
(\mathbf{n}_{so}^u)^{\!T}\Big(\Delta\mathbf{p}_s^{u} - h\,\mathbf{v}_o^u\Big)
\;\ge\;
-k\,h\, \Psi_{so}(\boldsymbol{\theta}^u),
\label{eq:sep_discrete}
\end{equation}
where $k>0$ is a gain. Then a linear inequality in $\Delta\boldsymbol{\theta}^{u}$ can be obtained:
\begin{equation}
(\mathbf{n}_{so}^u)^{\!T}\mathbf{J}_s(\boldsymbol{\theta}^u)\,\Delta\boldsymbol{\theta}^{u}
\;\ge\;
-k\,h\, \Psi_{so}(\boldsymbol{\theta}^u) + h\,(\mathbf{n}_{so}^u)^{\!T}\mathbf{v}_o^u.
\label{eq:ineq_delta_theta}
\end{equation}

Let $\Delta\boldsymbol{\theta}^{u}_{\mathrm{nom}}$ denote a nominal joint increment(without considering of constraints). For the $m$ active unilateral constraints, define the \emph{constraint row Jacobians}
\begin{equation}
\mathbf{H}_i^{T}
\;\triangleq\;
(\mathbf{n}_i^u)^{\!T}\mathbf{J}_{s_i}(\boldsymbol{\theta}^u)\in\mathbb{R}^{1\times n},
\qquad i=1,\dots,m,
\label{eq:H_i_def}
\end{equation}
and stack them into $\mathbf{H}\in\mathbb{R}^{m\times n}$.
Define the right-hand side residual
\begin{equation}
b_i
\;\triangleq\;
\Big(-k\,h\, \Psi_i(\boldsymbol{\theta}^u) + h\,(\mathbf{n}_i^u)^{\!T}\mathbf{v}_{o_i}^u\Big)
-\mathbf{H}_i^{T}\Delta\boldsymbol{\theta}^{u}_{\mathrm{nom}},
\qquad i=1,\dots,m,
\label{eq:b_i_def_proj}
\end{equation}
and let $\mathbf{b}=[b_1,\dots,b_m]^T$.
We seek a corrected increment in the normal-space form
\begin{equation}
\Delta\boldsymbol{\theta}^{u}
\;=\;
\Delta\boldsymbol{\theta}^{u}_{\mathrm{nom}} + \mathbf{H}^{T}\boldsymbol{\eta}^{u},
\qquad
\boldsymbol{\eta}^{u}\in\mathbb{R}^{m} \ge 0,
\label{eq:delta_theta_proj_eta}
\end{equation}

where $\boldsymbol{\eta}^{u}$ are nonnegative \emph{complementarity multipliers} associated with the unilateral separation constraints.
Substituting \eqref{eq:delta_theta_proj_eta} into the stacked inequalities
$\mathbf{H}\Delta\boldsymbol{\theta}^{u}\ge \mathbf{H}\Delta\boldsymbol{\theta}^{u}_{\mathrm{nom}}+\mathbf{b}$
yields
\begin{equation}
\mathbf{H}\mathbf{H}^{T}\boldsymbol{\eta}^{u} - \mathbf{b} \;\ge\; \mathbf{0}.
\label{eq:HH_eta_ge_b}
\end{equation}
Define the slack
\begin{equation}
\mathbf{w}^{u}
\;\triangleq\;
\mathbf{H}\mathbf{H}^{T}\boldsymbol{\eta}^{u} - \mathbf{b}\in\mathbb{R}^{m}.
\label{eq:w_def_proj}
\end{equation}
The complementarity conditions
\begin{equation}
\mathbf{0}\le \boldsymbol{\eta}^{u}\ \perp\ \mathbf{w}^{u}\ge \mathbf{0}
\label{eq:lcp_compact_proj}
\end{equation}
form the LCP
\begin{equation}
\mathbf{w}^{u} = \mathbf{M}\boldsymbol{\eta}^{u}+\mathbf{q},
\qquad
\mathbf{M}\triangleq \mathbf{H}\mathbf{H}^{T},
\quad
\mathbf{q}\triangleq -\mathbf{b}.
\label{eq:lcp_standard_proj}
\end{equation}

\eqref{eq:lcp_compact_proj}--\eqref{eq:lcp_standard_proj} is a special case of an MCP with bounds $\boldsymbol{\ell}=\mathbf{0}$ and $\boldsymbol{u}=+\infty$.

\begin{algorithm}[!t]
\caption{Rigid-body motion prediction}
\label{alg:mb_contact_interactions_loop}
\DontPrintSemicolon
\SetKwInOut{Input}{Input}
\SetKwInOut{Output}{Output}

{\footnotesize
\setlength{\intextsep}{4pt}
\setlength{\textfloatsep}{6pt}
\setlength{\floatsep}{4pt}
\setlength{\abovedisplayskip}{2pt}
\setlength{\belowdisplayskip}{2pt}
\setlength{\abovedisplayshortskip}{1pt}
\setlength{\belowdisplayshortskip}{1pt}
\renewcommand{\baselinestretch}{0.92}\selectfont

\SetAlgoSkip{2pt}
\SetAlgoInsideSkip{2pt}
\SetInd{0.8em}{0.8em}

\Input{Initial bodies $\{x_i^{0}=(\boldsymbol{q}_i^{0},\boldsymbol{\nu}_i^{0})\}_{i=1}^N$, initial tool state $tool^{0}$, timestep $h$, number of steps $T$, parameters}
\Output{Trajectories $\{x_i^{u}\}_{u=0}^{T}$ and $\{tool^u\}_{u=0}^{T}$}

\BlankLine
\For{$u \leftarrow 0$ \KwTo $T-1$}{

\noindent
\begin{tabular}{@{}p{0.49\linewidth}@{\hspace{0.02\linewidth}}p{0.49\linewidth}@{}}
\stepblock{1}{Predict (no contact):} &
\stepblock{2}{Detect contacts:} \\

$\{\boldsymbol{\nu}_{i,\mathrm{free}}^{u}\} \leftarrow \mathrm{Ext.Forces}$ \eqref{eq:free_velocity}; &
$\boldsymbol{a}_g^{u} \leftarrow \mathrm{Det.Ground}$ \eqref{non_penetration_disc}; \\

$tool_{\mathrm{free}}^{u} \leftarrow \mathrm{ToolCmd.}$; &
$\boldsymbol{a}_{tb}^{u} \leftarrow \mathrm{Det.ToolBody}$ \eqref{eq:two_body_nonpen}; \\

&
$\boldsymbol{a}_{bb}^{u} \leftarrow \mathrm{Det.BodyBody}$ \eqref{eq:two_body_nonpen}; \\
\end{tabular}

\vspace{2pt}

\noindent
\begin{tabular}{@{}p{0.49\linewidth}@{\hspace{0.02\linewidth}}p{0.49\linewidth}@{}}
\stepblock{3}{Solve contacts:} &
\stepblock{4}{Integrate poses:} \\

$\mathrm{SolveGroundMCP}$ \eqref{eq:free_velocity} - \eqref{eq:fj_impulse_4};&
$\{\boldsymbol{q}_i^{u+1}\} \leftarrow \mathrm{IntegrateCurrentPoses}$ \eqref{eq:config_update}; \\

$\mathrm{SolveToolBody\&BodyBody}$ \eqref{eq:free_velocity} - \eqref{eq:two_body_nonpen}; &
$ \{x_i^{t+1}\} \leftarrow \mathrm{Solve NextState}$; \\

\end{tabular}

} 

\Return{$\{x_i^{u}\}_{u=0}^{T},\ \{tool^u\}_{u=0}^{T}$}\;

} 
\end{algorithm}

\begin{algorithm}[!t]
\caption{Whole-body obstacle avoidance} 
\label{alg:LCP_part}
\DontPrintSemicolon
\SetKwInOut{Input}{Input}
\SetKwInOut{Output}{Output}
\Input{Joint trajectory $\boldsymbol{\theta}^{t}$, timestep $h$, number of steps $T$, parameters}
\Output{Increment of Joint Trajectory $\Delta\boldsymbol{\theta}^{t}_{\mathrm{nom}}$ }
\BlankLine
\For{$t \leftarrow 0$ \KwTo $T-1$}{
    $\Delta\boldsymbol{\theta}^{t}_{\mathrm{nom}} \leftarrow \mathrm{Solve LCP}$\eqref{eq:lcp_standard_proj};
}
\Return{$\{\boldsymbol{\theta_{\mathrm{cor}}}^{t}\}_{t=0}^{T} = \{\boldsymbol{\theta}^{t}\}_{t=0}^{T} + \{\Delta\boldsymbol{\theta}^{t}_{\mathrm{nom}}\}_{t=0}^{T}$ } ;
\end{algorithm}

\begin{algorithm}[!t]
\caption{Unified complementarity framework (Unicomp)}
\label{alg:whole_framework}
\DontPrintSemicolon
\SetKwInOut{Input}{Input}
\SetKwInOut{Output}{Output}

\Input{
Desired object waypoints $\{\boldsymbol{q}_{o}^{\star,t}\}_{t=1}^{T}$,
initial states $(\boldsymbol{q}_{o}^{0},\boldsymbol{\nu}_{o}^{0},\, \boldsymbol{\theta}^{0})$, parameters
}
\Output{
Robot joint trajectory $\{\boldsymbol{\theta}^{t}\}_{t=0}^{T}$
and object/tool trajectories
}

Initialize robot, rigid bodies and tool state;

\While{$t \le T$}{
    \textbf{1}. Generate desired tool command from current object waypoint \\
    \textbf{2}. Interaction solve (Alg. \ref{alg:mb_contact_interactions_loop}): advance tool--object states under contact \\
    \textbf{3}. In parallel: compute raw joint command that tracks the tool command (IK / impedance)~\cite{lynch2017modern}, and whole-body avoidance (Alg. \ref{alg:LCP_part}) \\
    \textbf{4}. Compose and apply joint update \\
}
\Return{$\{\boldsymbol{\theta}^{t}\}_{t=0}^{T}$}\;
\end{algorithm}

\input{results}

\input{benchmarks}

\begin{credits}
\end{credits}
%
%
%
\bibliographystyle{splncs04}
\bibliography{bibliography}
%




\end{document}

%% file: results.tex
\section{Numerical Simulations}

\subsection{Non-prehensile rigid-body interactions}
\label{sec:nonprehensile}

\noindent\textbf{Goal.}  
We demonstrate that the proposed formulation Unicomp captures \emph{intermittent contact} and \emph{stick--slip} transitions
without prescribing contact mode, as implemented in Algorithm~\ref{alg:mb_contact_interactions_loop}.

\noindent\textbf{Scenario and setups.}  
A rigid block of mass $m=0.8\,\mathrm{kg}$ moves on a horizontal plane with Coulomb friction coefficient $\mu=0.5$.
We evaluate the same contact indicators under two actuation modalities while keeping an identical contact model:
(a) \emph{Direct CoM forcing} applies a prescribed planar wrench at the block center of mass (check supplementary material); (b) \emph{Tool-driven pushing} uses a spherical tool (green marker) to push the block, producing a realistic, time-varying interaction force and coupled translation--rotation.
In the tool-driven case we additionally plot the tool force to indicate when pushing is active.

\noindent\textbf{Quantities shown in the plots.}  
Let $\boldsymbol{\Lambda}=[\Lambda_t\;\Lambda_o\;\Lambda_r]^\top$ be the friction impulses at the ECP,
$\Lambda_n$ the normal impulse, and $Q^{-1}\in\mathbb{R}^{3\times 3}$ the ellipsoidal limit-surface matrix.
We plot the normalized limit-surface coordinate
\begin{equation}
s \;\triangleq\; \frac{\boldsymbol{\Lambda}^\top Q^{-1}\boldsymbol{\Lambda}}{(\mu \Lambda_n)^2},
\label{eq:s_def}
\end{equation}
and its component-wise normalizations
\begin{equation}
\rho_t \;\triangleq\; \sqrt{\Big(\frac{\Lambda_t}{\mu \Lambda_n e_t}\Big)^2+\Big(\frac{\Lambda_o}{\mu \Lambda_n e_o}\Big)^2},\qquad
\rho_r \;\triangleq\; \Big|\frac{\Lambda_r}{\mu \Lambda_n e_r}\Big|,
\label{eq:rho_def}
\end{equation}
where $e_t,e_o,e_r$ are consistent with our parameterization of $Q^{-1}$.
In the common diagonal case $Q^{-1}=\mathrm{diag}(1/e_t^2,1/e_o^2,1/e_r^2)$, the normalization becomes
$s=\rho_t^2+\rho_r^2$, which separates tangential and torsional saturation directly. The scalar $s$ measures how close the friction impulse is to the normalized limit-surface boundary.
Under maximum dissipation, sliding solutions saturate on the boundary ($s\approx 1$); sticking stays strictly inside ($s<1$).
The pair $(\rho_t,\rho_r)$ further indicates whether saturation is dominated by tangential friction ($\rho_t$) or torsional friction ($\rho_r$).
The contact breaks as $\Lambda_n\to 0$.

\noindent\textbf{Automatic mode annotation in the figures.}  
To improve readability, we annotate each time interval as \emph{break/stick/slide} from the solver outputs using $(\Lambda_n,s)$ with small numerical thresholds:
\emph{break} if $\Lambda_n\le \varepsilon_n$; otherwise \emph{slide} if $s\ge 1-\varepsilon_s$ and \emph{stick} if $s<1-\varepsilon_s$.
These labels are \emph{post-processed} for visualization only (no hand-crafted mode switching), and are consistent with the theoretical signatures:
break $\Rightarrow \boldsymbol{\Lambda}\approx \mathbf{0}$, slide $\Rightarrow s\approx 1$, stick $\Rightarrow s<1$.
  
\begin{figure}[!t]
    \centering



        \includegraphics[width=0.9\linewidth]{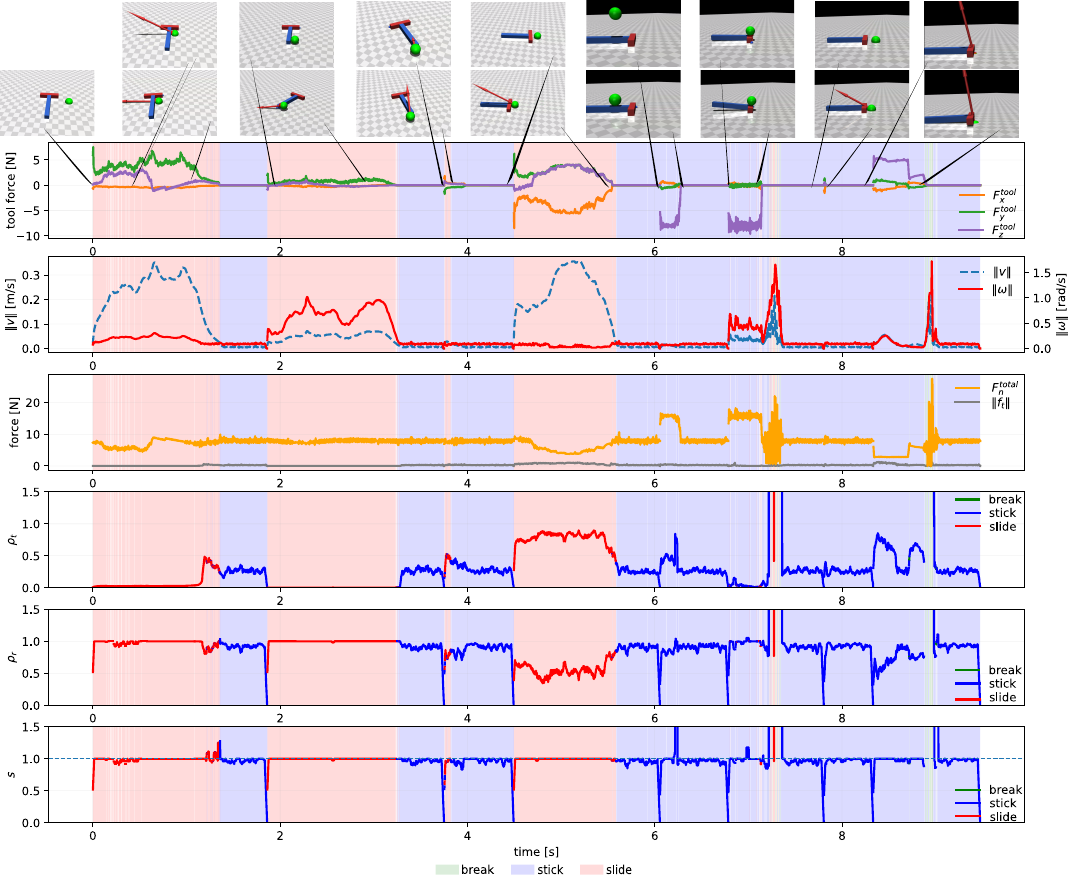}
        \caption{\textbf{Tool-driven pushing.} The tool force indicates when pushing is active, while $(\Lambda_n,\boldsymbol{\Lambda},s,\rho_t,\rho_r)$ are computed at the block--ground ECP. Despite non-smooth excitation from tool contact, the same complementarity signatures hold: $s\approx 1$ in sliding, $s<1$ in sticking, and impulses vanish when contact breaks.}
        \label{fig:nonprehensile_tool}
        \vspace{-0.3cm}

    \label{fig:nonprehensile_two}
\end{figure}
\FloatBarrier  
\noindent\textbf{Results.} Across both actuation modalities, the proposed formulation produces consistent contact signatures without prescribing discrete modes.
In the direct COM forcing case, the response cleanly reveals intermittent contact: as $\Lambda_n\to 0$, friction impulses vanish automatically, while sustained contact alternates between sticking ($s<1$) and sliding ($s\approx 1$).
In the tool-driven pushing case, the tool--object interaction introduces non-smooth, time-varying excitation, yet the same behavior persists:
sliding segments concentrate near the limit-surface boundary ($s\approx 1$), sticking remains strictly inside ($s<1$), and impulses disappear during contact loss.
Component-wise normalizations further clarify the sources of saturation, with $\rho_t$ capturing tangential friction saturation and $\rho_r$ capturing torsional saturation.

\subsection{Interactive planar pushing with non-convex contact}

Non-prehensile pushing becomes challenging when the supporting contact is non-convex or disconnected. In such cases, the active support region can switch abruptly (e.g., between different legs of a table), making point-contact or single-patch assumptions brittle without explicit contact-mode enumeration.

We perform interactive planar pushing using a spherical tool, shown in green, with a user-specified target position shown in red. The tool is driven by an impedance controller that tracks the target and generates contact forces upon interaction with the object. At each simulation step, the resulting tool–object wrench and object–ground contact dynamics are resolved using our time-stepping model. All experiments run at a fixed step size of 
$h=1ms$ (1000 Hz) on a single CPU core. For visualization, we record object center-of-mass (CoM) and equivalent contact point (ECP) trajectories.

\begin{figure}[!t]
    \centering
    \includegraphics[width=0.95\linewidth]{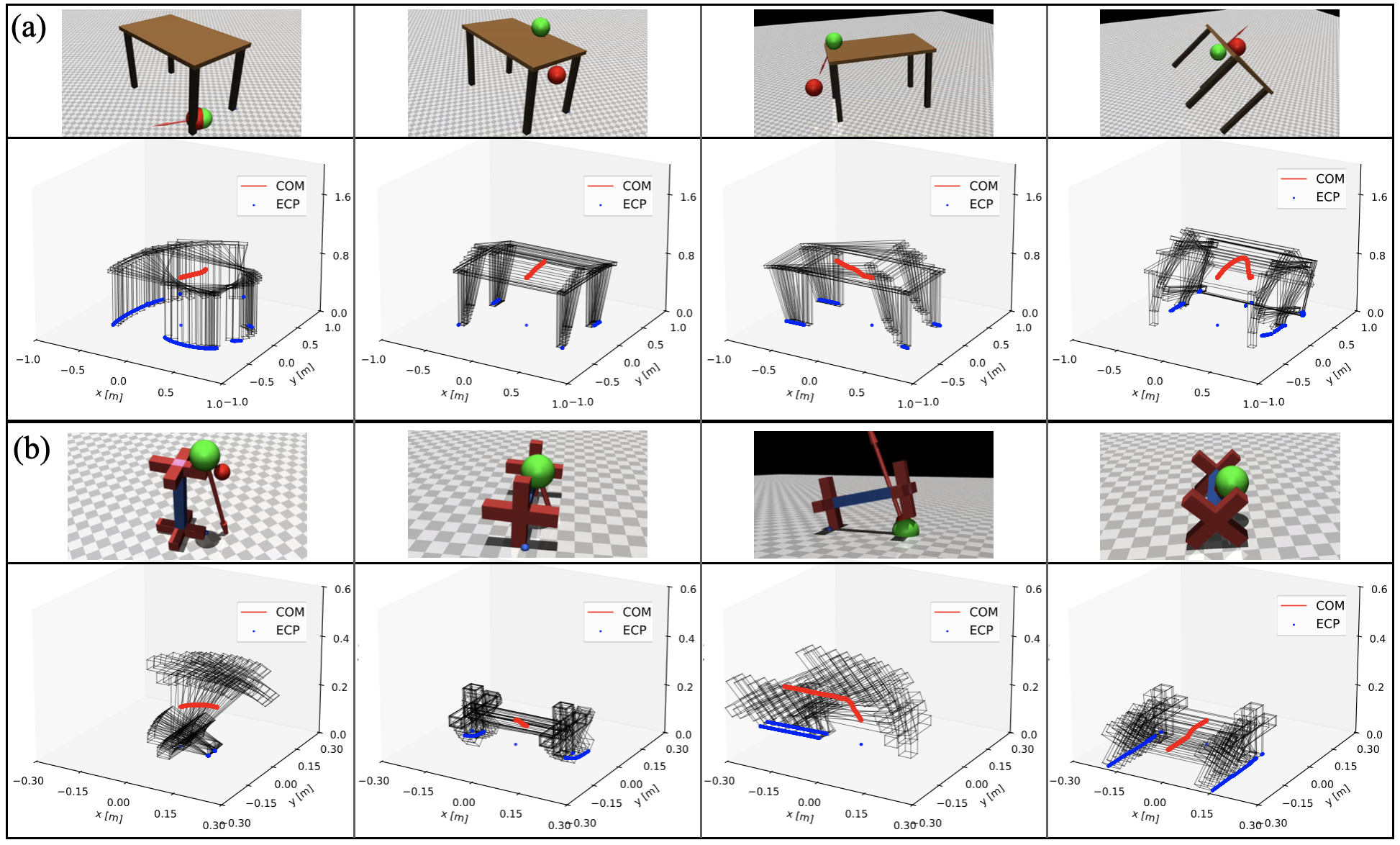}
    \caption{\textbf{Interactive planar pushing with non-convex support at $1000\,\mathrm{Hz}$.}
    The green sphere denotes the impedance-controlled tool and the red sphere the user-specified target.
    For each pushing action (column), we show a snapshot (top) and the corresponding CoM (red) and object--ground ECP (blue) trajectories with pose stacks (bottom).
    (a) Table-like object; (b) dumbbell-like object.}
    \label{fig:nonconvex_pushing}
    \vspace{-0.3cm}
\end{figure}

Figure~\ref{fig:nonconvex_pushing}(a) shows pushing a table-like object with multiple separated support contacts. As the object moves, the effective support shifts between different legs, reflected by the ECP migrating across distinct ground regions, while the CoM trajectory remains smooth. Across both non-convex objects, unicomp produces physically consistent motion and handles support switching and intermittent contact at interactive rates without explicit contact-mode enumeration.

\subsection{Comparison with Mujoco}

\begin{figure}[!t]
    \centering
    \includegraphics[width=0.89\linewidth]{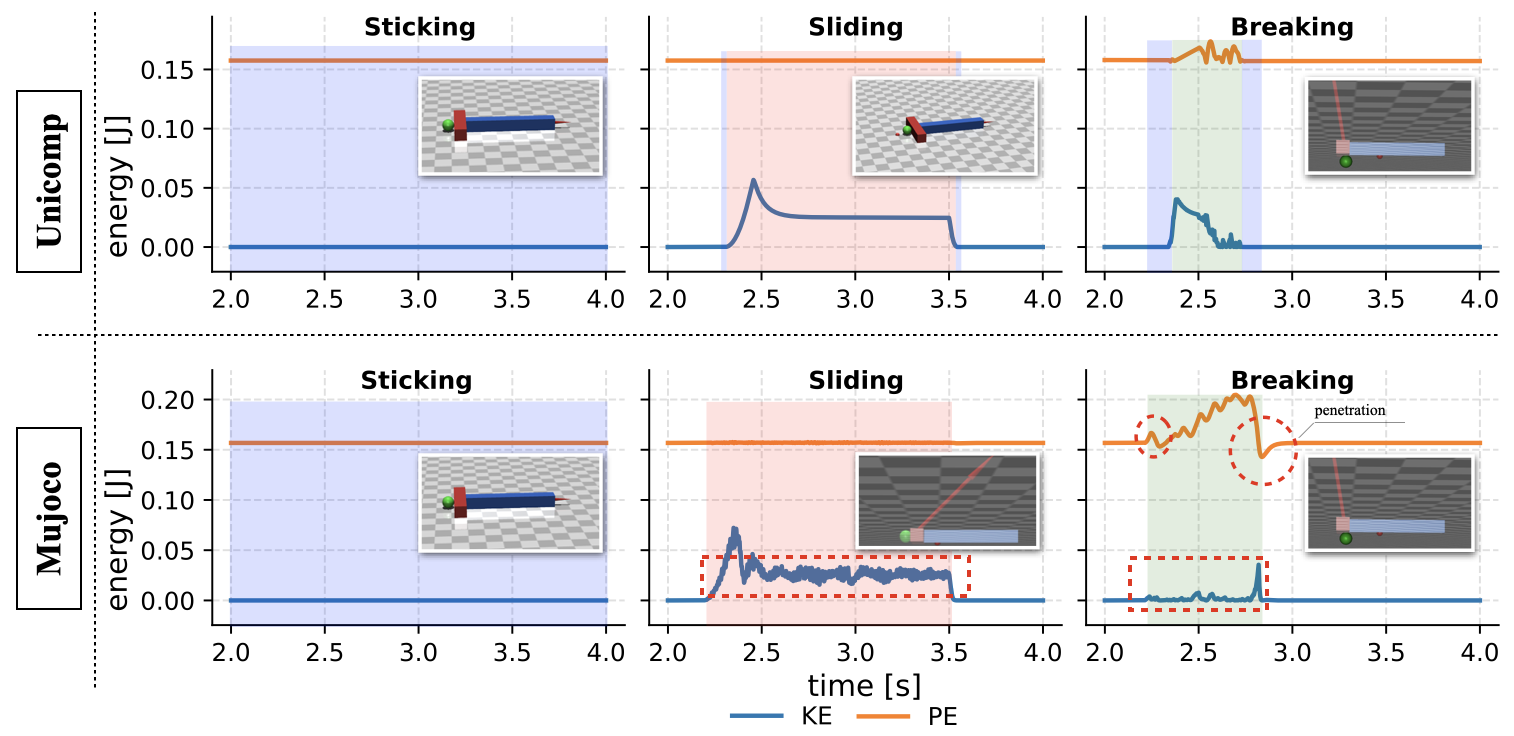}
    \caption{Comparison of Unicomp with Mujoco. During sliding and breaking modes, phantom forces arise in Mujoco, whereas our plots remain stable and smooth.}
    \label{fig:compare_mujoco}
    \vspace{-0.5cm}
\end{figure}

We design a comparative experiment with MuJoCo \cite{todorov2012mujoco}, using the same impedance controller to drive a small sphere (mass $0.0335\,\mathrm{kg}$) into a non-convex rigid body (mass $0.8\,\mathrm{kg}$).
Fig.~\ref{fig:compare_mujoco} summarizes the results: our simulation (Unicomp) is shown in the upper row and MuJoCo in the lower row.
The blue background indicates the sticking phase, red indicates sliding, and green indicates breaking.
The yellow curve denotes the block's gravitational potential energy, while the blue curve denotes its kinetic energy.

\textbf{Scene 1 (sticking).} With a maximum actuation of $1\,\mathrm{N}$ and target position $\big[0.0,\,0.35,\,0.02\big]$, the sphere cannot push the block. As shown in Fig.~\ref{fig:compare_mujoco}, our results qualitatively match MuJoCo, exhibiting similar sticking behavior and energy evolution.

\textbf{Scene 2 (sliding).} Increasing the command to $10\,\mathrm{N}$ produces sufficient normal force to induce sustained sliding. Fig. \ref{fig:compare_mujoco} shows smoother trajectories than MuJoCo, which we attribute to representing the distributed patch contact via an ECP. In the highlighted region, MuJoCo exhibits force fluctuations that slightly deviate the block’s motion from the nominal collision direction.

\textbf{Scene 3 (breaking).} An oblique impact generates a lifting wrench that transitions the ground contact away from full patch contact. Our model captures this evolution and the resulting sticking-to-sliding transition during separation (Fig. \ref{fig:compare_mujoco}). In contrast, MuJoCo shows repeated ground penetration, injecting non-physical energy, highlighted by the rapid kinetic energy increase.

%% file: benchmarks.tex
\section{System-Level Evaluation}

\label{sec:multi_body_contact_propagation}

\begin{figure}[!t]
    \centering
    \includegraphics[width=0.8\linewidth]{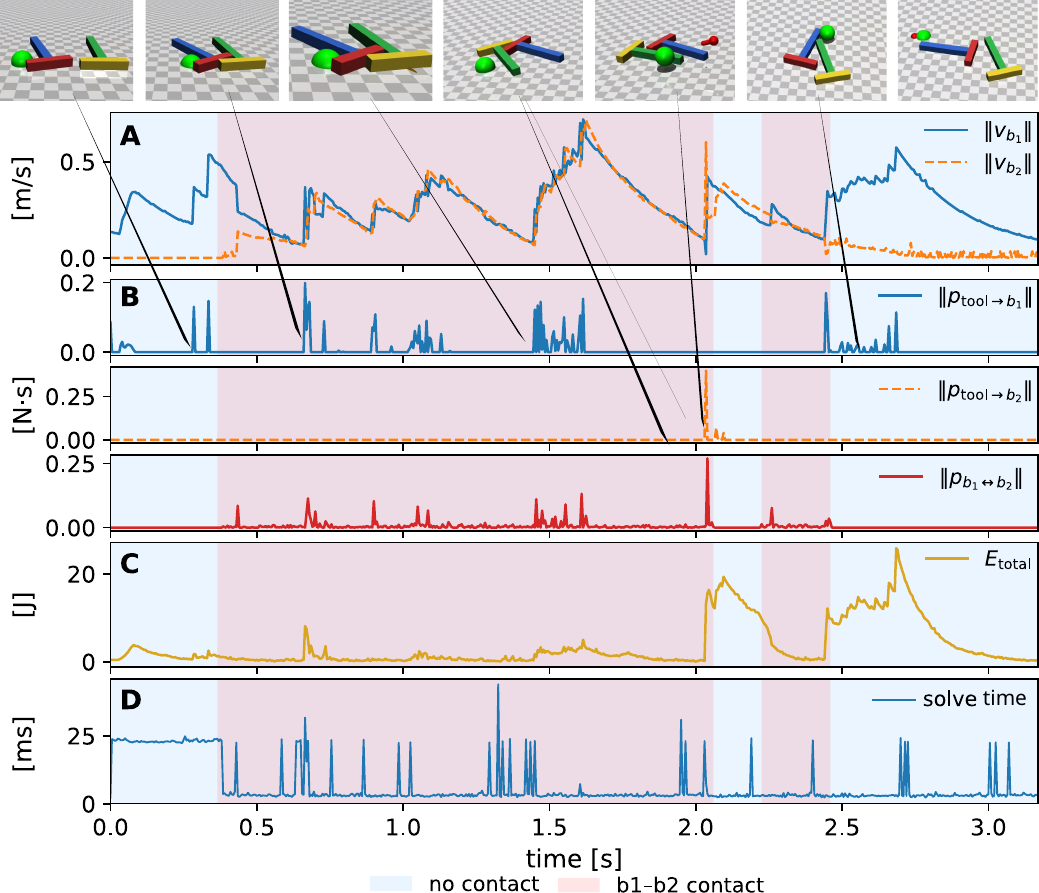}
    \caption{Energy and computation profiles for multi-object interaction.}
    \label{fig:placeholder}
    \vspace{-0.8cm}
\end{figure}
\textbf{Multi-Body Contact Propagation.} We demonstrate impulse and motion propagation in a multi-body scene where a tool impact on one block induces a secondary interaction with another (tool $\rightarrow b_1 \rightarrow b_2$). This example shows that coupled contact effects emerge without explicit enumeration of contact modes such as impact–stick or impact–slide. At each time step, the system is updated using a modular contact pipeline. Block–ground contact is resolved using a complementarity-based normal constraint with dissipative friction. Tool–block and block–block interactions are then handled via scalar normal complementarity conditions and tangential impulses limited by Coulomb friction, applied as equal-and-opposite point impulses. These updates are coupled within a single time step through sequential impulse application, yielding a robust and efficient operator-splitting scheme. We log linear and angular velocities of the tool and both blocks, along with normal and tangential impulses. We report propagation delay between tool–$b_1$ and $b_1$–$b_2$ impulses and peak velocity transfer ratios. The time series shows a clear propagation pattern: the tool–$b_1$ impulse increases $b_1$ velocity, followed by activation of the $b_1$–$b_2$ impulse and acceleration of $b_2$. Friction and ground contact bound the final motion, confirming that our pipeline captures emergent multi-body propagation effects relevant to planning and execution in cluttered environments.

\textbf{Planar pushing with non-convex contact patch.}
We implement a light planar pushing planner that drives the object to a desired planar pose
$g^\star \in SE(2)$ in closed loop. Let the planar state be $x_k = (p_k,\theta_k)$ with $p_k\in\mathbb{R}^2$.
At each control step, we generate candidate pushing actions by \emph{surface-sampling-based contact selection}
(i.e., discretizing side surfaces into contact samples with outward normals). For each candidate action $u_k$
(contact sample index + tool motion direction/magnitude), we evaluate a one-step prediction using our
complementarity-based dynamics model,
$x_{k+1} = f(x_k, u_k),$
and select the action that best reduces the goal error via a cost,
$u_k^\star = \arg\min_{u_k \in \mathcal{U}(x_k)} 
\Bigl(
\|p_{k+1}-p^\star\|_2^2 + w_\theta\,\mathrm{wrap}(\theta_{k+1}-\theta^\star)^2
\Bigr),
\label{eq:one_step_push_cost}$
where $w_\theta$ balances translation and yaw alignment, and $\mathcal{U}(x_k)$ denotes feasible candidates. The chosen contact sample defines a reference tool configuration for (i) a collision-free approach to a safe offset
along the outward normal and (ii) a short push segment that realizes the selected action. If contact breaks or the
cost reduction stalls, the planner re-selects a surface sample and repeats the approach--push cycle.
The current planner is intentionally simple and computationally efficient for real-time replanning. In future work, the same predictive model can be extended to finite-horizon MPC/trajectory optimization by optimizing
$\{u_k\}_{k=0}^{H-1}$ over a receding horizon.

\begin{figure}[t]
  \centering
    \includegraphics[width=0.93\linewidth]{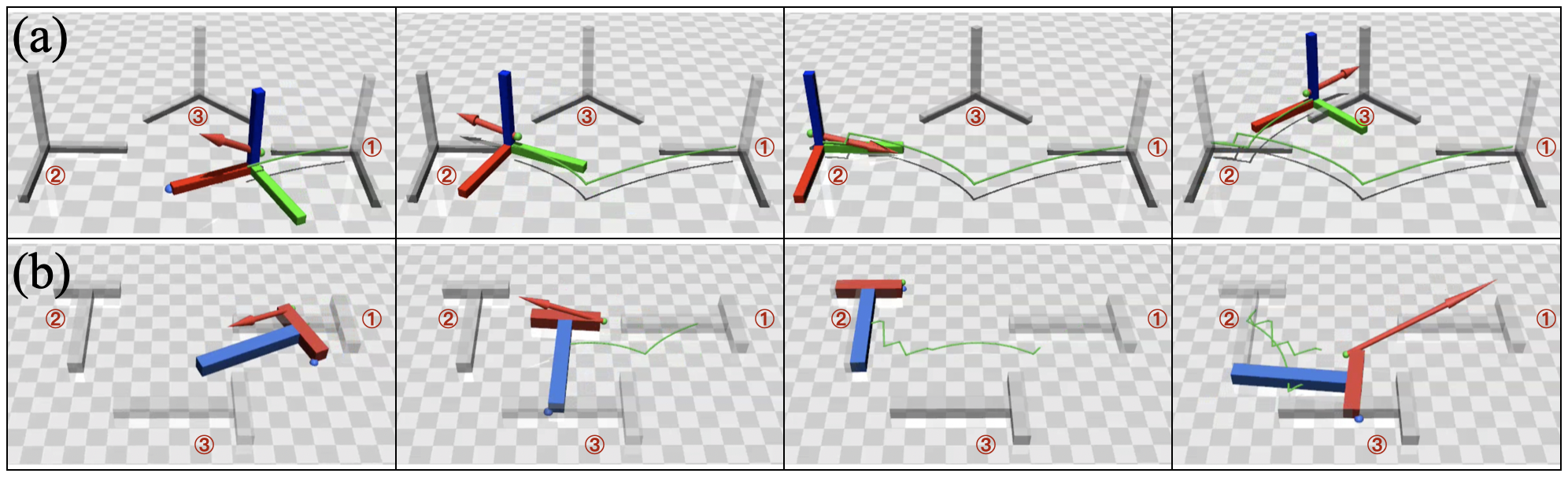}
  \caption{Planar pushing planner examples. (a) A frame-like object is pushed to different desired planar poses.
  (b) A cuboid is pushed to different desired poses.}
  \label{fig:planar_planner_examples}
  \vspace{-0.5cm}
\end{figure}

\textbf{Whole-Body Obstacle-Aware Pushing.}
\label{sec:whole_body_obstacle_aware_pushing}
We extend the planar pushing pipeline by introducing a robot arm and
using the end-effector (EE) to replace the previously free rigid tool. The planar planner provides step-wise
EE references (approach--push) on the same discrete-time grid used throughout the paper.
Let $x^{\mathrm{des}}_u$ denote the desired EE task-space reference produced by the planar planner at step $u$.
We employ a task-space impedance law to compute a nominal joint increment
$\Delta\boldsymbol{\theta}^{u}_{\mathrm{nom}}$ that tracks $x^{\mathrm{des}}_u$ over one step:
$\boldsymbol{\theta}^{u+1} = \boldsymbol{\theta}^{u} + \Delta\boldsymbol{\theta}^{u}$ ,
$
\Delta\boldsymbol{\theta}^{u}_{\mathrm{nom}} \triangleq h\dot{\boldsymbol{\theta}}^{u}_{\mathrm{imp}},
\label{eq:nominal_joint_increment}$
where $\dot{\boldsymbol{\theta}}^{u}_{\mathrm{imp}}$ is obtained from the impedance controller (e.g., via a Jacobian-based
mapping). In the absence of collision constraints, we would execute $\Delta\boldsymbol{\theta}^{u}=\Delta\boldsymbol{\theta}^{u}_{\mathrm{nom}}$. We approximate the robot geometry by a set of link-attached spheres and impose
the one-step separation constraints \eqref{eq:ineq_delta_theta} for all active sphere pairs. Using the stacked constraint
matrix $\mathbf{H}$ and residual $\mathbf{b}$ (Eqs.~\eqref{eq:H_i_def}--\eqref{eq:b_i_def_proj}), we correct the nominal
increment through the normal-space form
$\Delta\boldsymbol{\theta}^{u}
=
\Delta\boldsymbol{\theta}^{u}_{\mathrm{nom}} + \mathbf{H}^{T}\boldsymbol{\eta}^{u},
\boldsymbol{\eta}^{u}\ge 0,
\label{eq:wb_corrected_increment}$
where $\boldsymbol{\eta}^{u}$ is obtained by solving the LCP
\eqref{eq:lcp_compact_proj}--\eqref{eq:lcp_standard_proj},(Algorithm~\ref{alg:LCP_part}). This obstacle-aware correction is integrated into the overall Unicomp pipeline (Algorithm~\ref{alg:whole_framework}). Intuitively, when sufficient clearance exists the solution
yields $\boldsymbol{\eta}^{u}\approx 0$ and the arm tracks the impedance motion; near the safety boundary, the Complementarity multipliers activate and modify the motion.

\section{Conclusion}
We presented Unicomp, a discrete-time framework that models free-space motion and frictional contact using a single complementarity-based formulation. By coupling linear and nonlinear complementarity problems, the approach enables consistent contact-mode transitions without fixed-contact assumptions. For planar patch contact, we introduced an ellipsoidal limit-surface model derived from the maximum power dissipation principle, capturing coupled force–moment effects while remaining suitable for real-time optimization. Experimental results demonstrate stable and physically consistent behavior across a range of contact-rich manipulation tasks, supporting Unicomp as a practical foundation for robust manipulation planning in unstructured environments.